# Learning Hands-On Electronics from Home: A Simulator for Fritzing

Andres Faiña (✉)

Robotics, Evolution and Art Lab (REAL),
IT University of Copenhagen,
Copenhagen S 2300, Denmark
anfv@itu.dk

**Abstract.** The recent pandemic has forced us to teach online, which is especially difficult for hand-on courses in robotics, like basic electronics. In this paper, I present a simulator which tries to replicate the same experience students will encounter during the exercises in the laboratory. The simulator has been developed in Fritzing, uses realistic multimeters for measuring and checks for common mistakes. Results show not only that the simulator was extremely useful during the pandemic, but also that it can supplement laboratory exercises when teaching in-classroom.

**Keywords:** Learning electronics, Electronics simulator, Fritzing, Educational tools

## 1 Introduction

Robotics is an interdisciplinary area which overlaps with the fields of mechanical engineering, electronics, and computer science. A basic understanding of these three fields is necessary to have a global understanding of robotics. However, learning to design and build electronic circuits is a difficult task for students without an engineering background. Students need to learn the basic principles of electric circuits which include electronic units, basic rules of circuits and their representation in schematics. Nevertheless, this knowledge is not enough to implement them.

To implement circuits, it is also necessary to learn practical skills that cannot be taught with just a schematic circuit. For example, a breadboard is often used to implement prototypes, and printed circuit boards (PCBs) are used for production. New electronic practitioners often struggle to build implementations of schematic circuits [1,2]. This happens as the schematic view is a symbolic representation of the circuit while circuits made on protoboards or PCBs are implementations with their own specific particularities. Moreover, debugging a circuit after implementing it on a breadboard or a PCB is usually necessary and requires knowing how to use multimeters and oscilloscopes. Therefore, implementation and debugging skills need to be learned as well.

The traditionally approach to teach electronics is to combine theoretical explanations about circuits using schematics with a practical session where students implement circuits in breadboards [3,4]. And after they become more experienced, they learn how to produce PCBs [5]. However, practical sessions are not always possible as they require (1) that the students have the electronic components and hardware tools and (2) that a teacher is available during the practical sessions.

During the current pandemic, a lot of electronic courses were taught online, and students did not have access to these practical sessions. In my case, I teach the first course of the robotic specialization, which is called "How to Make (Almost) Anything", where M.Sc. Computer Science students without engineering background learn mechanics, electronics and microcontroller programming by prototyping robots and machines. There are only four lectures about electronics, but the students should hand in a functional prototype. Thus, during the pandemic, the only solution was to use an electronic simulator to teach both the theory and the practicalities of implementing electronic circuits.

There are several electronic simulators available, but most of them focus on only one view of the circuit: the advanced simulators focus on schematics and simulators for learners use a breadboard (and wiring) view. In addition, few of them provide feedback on important mistakes: short-circuit detection, voltage, current or power exceeded, etc. Finally, debugging tools are often simplified and do not represent real tools accurately. For example, some simulators allow to measure a voltage with just one probe (the other one is assumed to be the ground terminal), which obfuscates that voltage is a relative measurement. Other simulators use specific parts for voltmeters, ammeters and ohmmeters, while in a real-life scenario it is standard to use a multimeter. Using this tool, students need to change the mode of the multimeter and connect its probes correctly to be able to measure. In order to overcome these limitations and provide a "practical" session with a simulator as close to reality as possible, I implemented a simulator in Fritzing, an open-source electronic design automation program.

Table 1: Comparison of different electronic circuit simulators.

| Simulator | View | | Simulation | | | | | | | |
|---|---|---|---|---|---|---|---|---|---|---|
| | Breadboard | Schematic | DC | AC | Arduino | Real Multimeters | Failure Detection | Create Components | Platform | License |
| Fritzing | x | x | x | -[a] | - | x | x | x | All[b] | GPL |
| TinkerCAD | x | - | x | x | x | - | Few[c] | - | Web | Closed |
| KiCad | - | x | - | x | - | - | - | x | All[b] | GPL |
| CircuitLab | - | x | x | x | - | - | - | x | Web | Closed |
| VirtualBreadboard | x | - | x | - | x | - | - | - | Win | Closed |
| SimulIDE | x[d] | | x | x | x | - | -[e] | x | All[b] | GPL |
| Wokwi | x | - | - | - | x | - | - | x | Web | MIT |
| Proteus VSM | - | x | x | x | x | - | - | x | Win | Closed |

[a] It will be implemented in future versions.
[b] Windows, MacOs and Linux.
[c] Polarity in electrolytic capacitors and, in some items, maximum voltage and maximum current.
[d] Mixes schematic symbols and images of Arduinos and sensors.
[e] Only for current in LEDs.

## 2 Related Work

One of the most popular electronic simulators is SPICE [6] and its different forks such as LTspice [7] and Ngspice [8]. These simulators take the definition of an electronic circuit through a netlist and can perform different types of analyses: operating point, transient, noise, etc. They simulate electronic circuits accurately, but they are difficult to use for beginners. For example, SPICE and Ngspice do not offer a schematic view of the circuit. Ngspice can be used inside KiCad [9], a popular suite for electronic design automation, which provides a schematic editor and basic interaction with the simulator. However, it only has a very limited set of parts that can be simulated as SPICE models usually have proprietary licenses. Thus, users need to select a specific component, find its SPICE model and add it to the simulator. In contrast, LTspice comes with a large set of specific devices to simulate.

There are several simulators targeted to electronic learners such as TinkerCAD [10], SimulIDE [11] and WokWi[12]. These simulators are more oriented towards simulation of microcontrollers with limited capabilities of analog simulation. They provide a rich set of generic components, and the user can simulate them in a few clicks: drag and drop components, draw the wires and press run. However, they do not offer realistic multimeters and most of them cannot detect common failures: e.g., short-circuits or wrong polarity applied to electrolytic capacitors.

The most similar work to the simulator presented here is probably TinkerCAD, which offers some failure detections. However, the fact that it is closed source and does not allow creating new components makes it less flexible. In addition, it only provides a breadboard view, which does not encourage students to learn how to read and draw schematics.

There are plenty of electronic simulators and each of them has its own features. I have summarized the main features of the most popular ones in Table 1.

The simulator introduced in this paper is implemented inside Fritzing [13]. Fritzing is an electronic design automation software, which provides three concurrent views of the same circuit: breadboard, schematic and PCB. It is targeted to people without previous electronic knowledge, and it was developed to be easy-to use. Spice models are optional in Fritzing´s part format and Fritzing can export a spice netlist. However, there were extremely few components with spice models (basically, only resistors, inductors and capacitors).

## 3 Simulator

The following section describes the main features of the simulator and its implementation. The simulator has extended Fritzing by adding simulation capabilities.[1] Every time the simulator is triggered, a spice netlist is generated and simulated in a background thread using Ngspice [8]. Using the results of the spice simulation, the graphical user interface of Fritzing is updated.

The simulator tries to replicate what a student would observe when working with real hardware. Thus, it is not possible to measure anything without using a realistic multimeter. In this first implementation, the simulator only performs a direct current (DC) operation point analysis, which is enough for teaching basic electronic concepts. The simulator checks for failures and, if there are any, a smoke symbol is drawn on top of the damaged component.

[1]The simulator code is available at https://github.com/failiz/fritzing-app/tree/simulator2. The simulator code is being reviewed and it will be merged in the official Fritzing repository soon. Fritzing code is open source, but currently official binaries cost €8. The payments allow hiring a maintainer and developers. In addition, the simulator needs components with spice models, which are already merged in the Fritzing-parts repository: https://github.com/fritzing/fritzing-parts

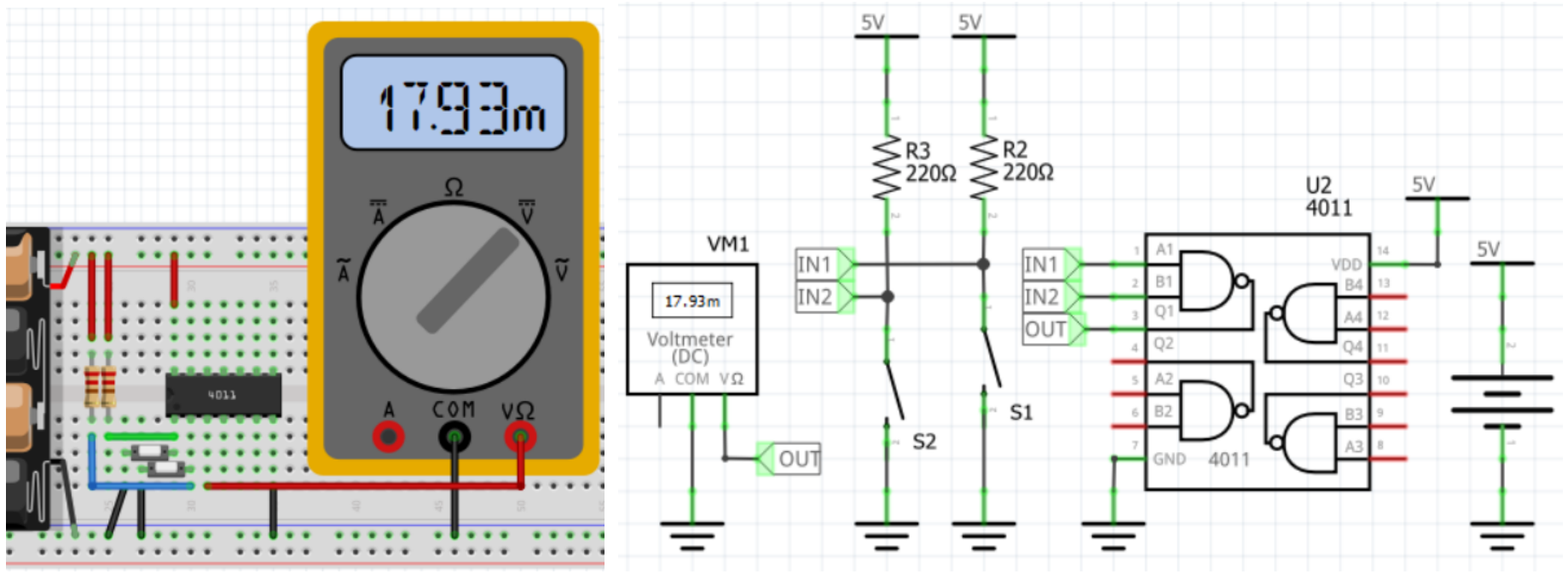

Fig. 1: The simulator allows students to use the breadboard view (left) or schematics view (right). In both views, a NAND gate (CD4011) is being tested. Two switches allow changing the input voltages and the output is measured with a multimeter. In the case shown, both inputs are 5V and the voltage at the output is 17.93mV.

The simulator comes disabled by default as it is still a beta feature in Fritzing, but it can be enabled in the properties menu. After that, a new button appears in the graphical user interface where the simulator can be started or stopped. Once the simulator has been started, the simulator will trigger a simulation each time that the user changes a property on one of the components (e.g., change the value of a resistor) or the wiring changes.

The main features of the simulator are described below.

**Multi view** Fritzing has three different views (breadboard, schematics and PCB) that are synchronized: changes in one view are propagated to the other views. E.g., when adding a component, the component is added in all the views, and when adding a wire, ratsnests are generated in the other views. Thus, the students can use any of them to build a circuit and simulate it. This allows students to build the circuit in the view they feel more comfortable. Once a circuit has been built in a view, ratsnest lines help to wire other views. Currently, multimeters cannot be connected and LEDs do not light up in the PCB view. However, breadboard and schematic views are fully functional, see Figure 1.

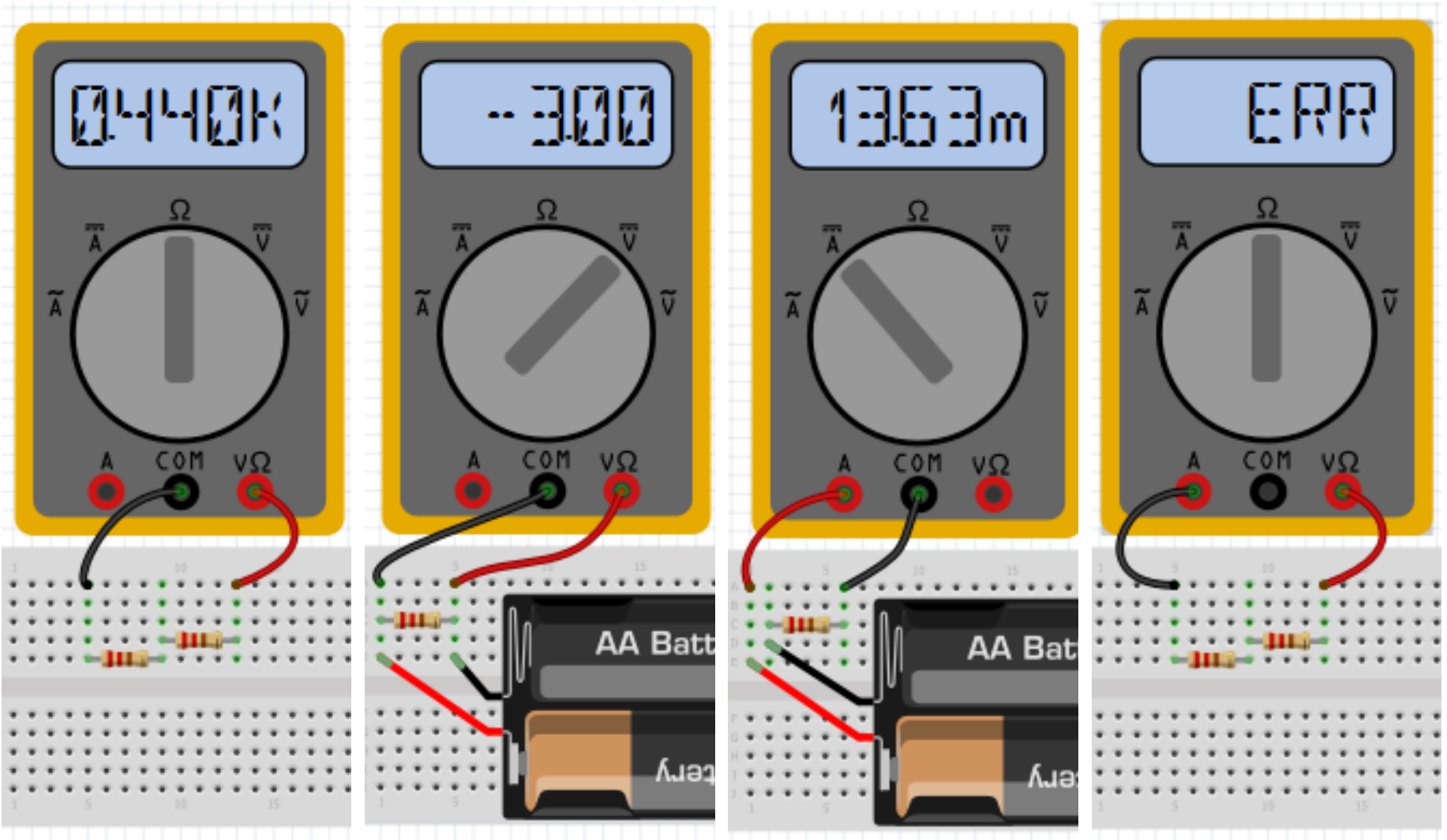

(a) Ohmmeter (b) Voltmeter (c) Ammeter (d) Bad wiring

Fig. 2: The realistic multimeters measuring different properties: (a) resistance, (b) voltage and (c) current. If a user connects the probes incorrectly, an error is reported. In (d), the multimeter is set to measure a resistance but the current probe ("A") is used instead of the common probe ("COM").

**Multimeters** A realistic multimeter part has been designed with five different variations: a voltmeter (DC and AC), an ammeter (DC and AC), and an ohmmeter. The user can change the type of the multimeter from the user interface (inspector pane). As most real multimeters, the part has three different connectors for the probes: a common one (COM), one for measuring voltages and resistances (VΩ) and one for measuring currents (A). The user needs to connect the probes to the points to measure and change the wiring of the circuit to read currents.

Once the simulation is performed, the multimeter will show the value measured in its screen. If the probes connected to the multimeter do not match with the measuring mode of the multimeter, an error message is shown on the screen (ERR). Figure 2 shows three different multimeters measuring a resistance, a DC voltage and a DC current, and a fourth one where the probes are not connected correctly.

**Failure detection** The simulator performs several checks to analyse if any of the components of the circuit is working outside its specifications. If so, a smoke image is added on top of the component to indicate it, see Figure 3. Currently, the simulator performs the following checks:

- Short circuit in batteries and power sources
- Max current (diodes and LEDs, inductors, IO pins)
- Max power (resistors)
- Reverse voltage (tantalum and electrolytic capacitors, IR sensors)
- Max voltage (capacitors, motors, IR sensors)

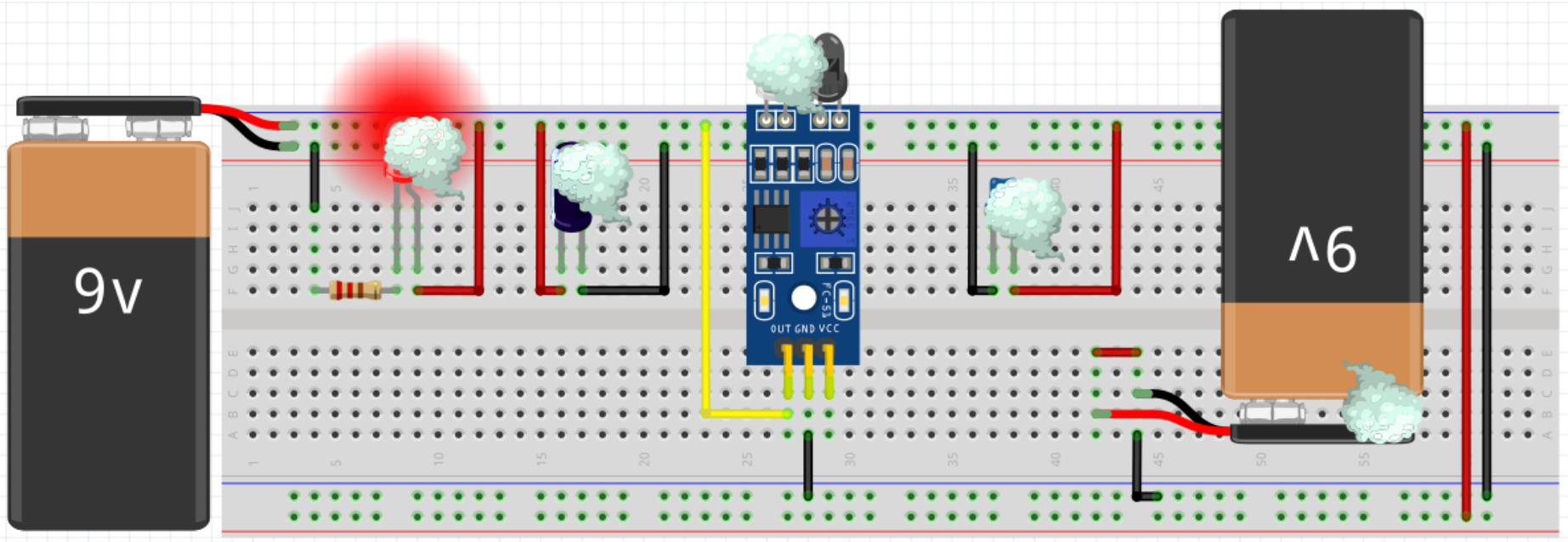

Fig. 3: Failure detection examples: The simulator checks for current exceeded (Led), reverse polarity (electrolytic capacitor), excess current in output pins (IR sensor), max voltage exceed (ceramic capacitor), and short-circuits (right battery).

**Components** The parts need to have a spice model in order to simulate them. Fritzing comes with around one thousand parts, but only a few of them have spice models. Thus, parts that can and cannot be simulated can coexist in the same sketch. To avoid confusions, all the parts that cannot be simulated are greyed out when the simulator is active. In addition, the spice model is shown in the inspector pane when a part is selected. To find parts that can be simulated, a new extra bin called ”Sim” has been created. In this bin, all the parts contain spice models and can be simulated.

The components represent generic versions of the most usual devices available at the laboratory, and they do not model a specific part. This simplifies the selection of components for the students, but it does not allow users to simulate circuits accurately. The spice model of a component cannot be changed through the user interface, but a spice model can use properties of the component. For example, the spice model of a potentiometer uses the maximum resistance property of the potentiometer, and this resistance can be changed by the user through the inspector pane. Advanced users can create new parts with specific spice models if they require accurate simulations.

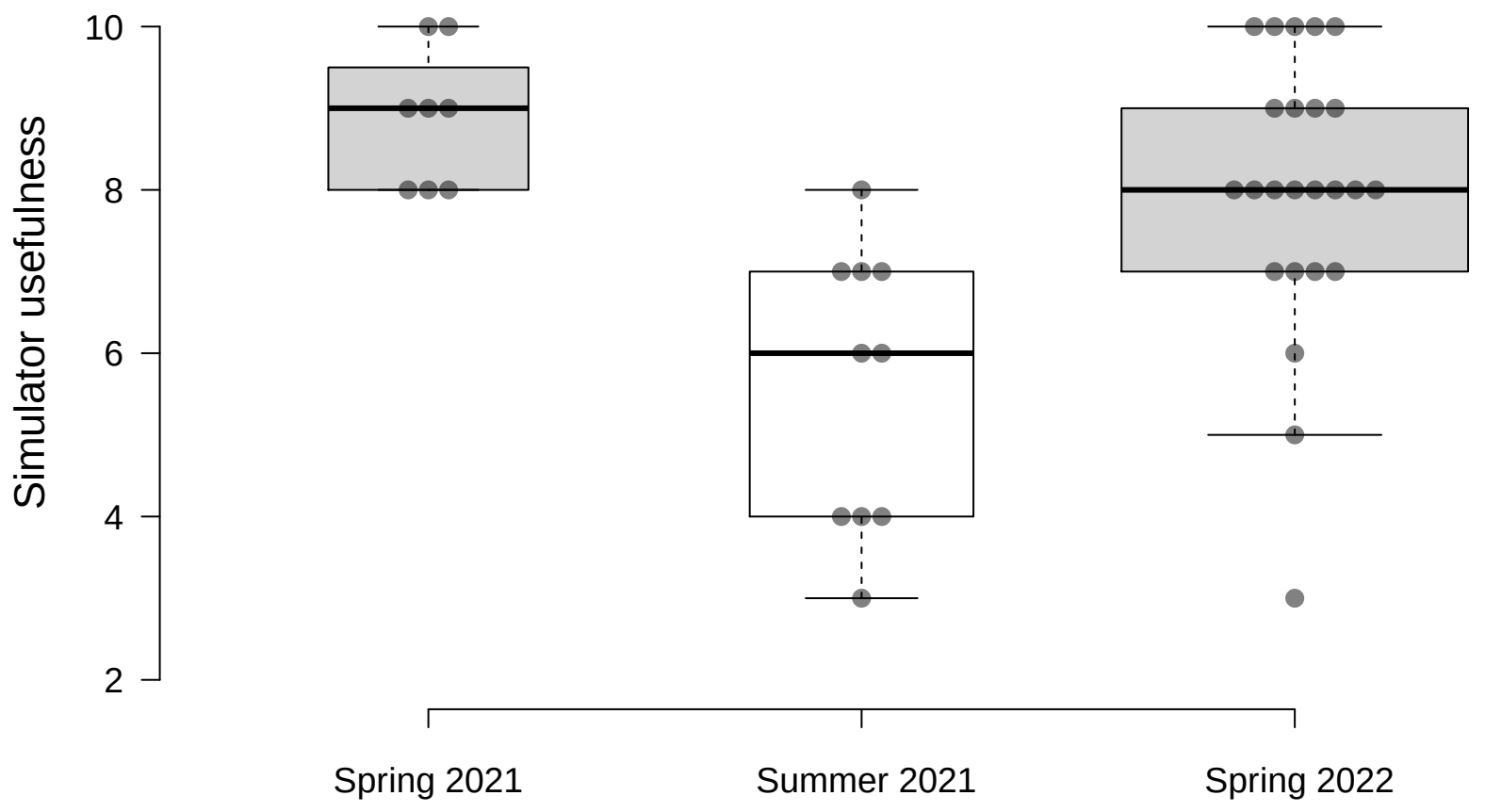

Fig. 4: Students graded the simulator usefulness from 0 to 10 during the course evaluation. Center lines show the medians; box limits indicate the 25th and 75th percentiles as determined by R software; whiskers extend 1.5 times the interquartile range from the 25th and 75th percentiles, outliers are represented by dots; data points are plotted as open circles. n = 8, 10, 24 sample points.

Currently, the components that can be simulated are resistors, capacitors, inductors, batteries and power supplies, potentiometers, switches, diodes and LEDs, transistors (bipolar and MOSFET), IR sensors (analogue and digital), DC motors and NAND gates.

## 4 Teaching with the Simulator

The simulator has been used in three classes of the course ”How to Make (Almost) Anything” at the IT University of Copenhagen in 2021 and 2022. The Spring 2021 course was taught completely online, and the Summer 2021 and Spring 2022 courses were taught physically at the university. In the 2021 courses, students used a preliminary (*alpha*) version of the simulator. The *alpha* version did not show the spice model in the user interface and did not have a simulation bin with all the parts that can be simulated. In addition, the simulation had to be triggered manually by pressing a button after any change in the circuit or in the properties of the components. Furthermore, only Windows and Linux versions were available and Mac users had to use a virtual machine. In the 2022 course, students used the latest version of the simulator and there was a binary for Mac users. Unfortunately, the Mac binary did not include a library and it was necessary to install the library in the system through the command line to be able to run the simulator.

In the lectures, students learned basic electronics, saw some examples and had a brief introduction to Fritzing and its simulator. During the following exercise session, students had to work on some mandatory assignments. These assignments consisted on building circuits with switches, potentiometers, IR sensors, transistors and DC motors. The students had to hand in the circuit assembled in the breadboard and their schematics. In courses taught at the university, students also had to build the circuits in the breadboard.

After each course, an anonymous and online survey about the simulator was carried out. The students were asked to answer ”Was the Fritzing simulator useful?” and they could grade it from 0 to 10, where 0 meant ”No, it was not useful” and 10 meant ”Yes, it helped me to learn electronics”. After the grading, they also could write in a text box what worked well in Fritzing and what could be improved. The survey was answered by 8 out of 30 students, 10 out of 18 and 24 out of 60 in the Spring 2021, Summer 2021 and Spring 2022 courses respectively.

Looking at the results of the survey, Figure 4, we can say that the simulator helps them to learn electronics. However, we can observe a big difference between the Spring and Summer courses in 2021. In the Summer course survey, I found three comments from students who reported that it was not useful as they could not use it in their computers (MacOS users), which probably explains the three lowest scores. Even discarding these three low scores, the simulator seems less useful when the students have access to practical exercises. In the Spring 2022 course, the simulator improvements have raised its usefulness, but it is still slightly below the Spring 2021 course.

I classified the answers of what could be improved in the Spring 2022 course (the one that used the latest version of the simulator). Five votes were for better support for MacOS, four votes for more parts with simulation models, and two votes for making the brightness of the LEDs more noticeable, simulate Arduinos, and adding interactive components (quickly change state of buttons and potentiometers). Finally, two persons also mentioned that working in both views, schematic and breadboard, at the same time could be confusing if unintentional connections are made.

In the surveys, I found some comments on the ability of the simulator to help them to learn electronics. Only one of them is negative ("*Fritzing was very helpful with creating schematics. Other than that, it was just easier to look at the real life electronics.*"). The rest are positive and can be classified in (1) better understanding of electronics (e.g., "*The simulation aspect of Fritzing was very nice. It was a good way of learning about switches and H-bridges and resistors, which can be quite abstract to learn.*" or "*I really liked how it helped me gain a good understanding about basic electronics.*"), (2) reducing stress of breaking components (e.g., "*I really enjoyed the simulation feature because it allowed me to be more confident in my circuits. I was very worried about damaging components so I liked to draw everything in Fritzing first just to make sure I wasn't gonna blow up any capacitors or the likes.*" or "*The parts that were there worked pretty well, and I liked being able to simulate circuits and getting some kind of security before actually building it in real life.*") and (3) helping to translate schematics into implementations (e.g., "*The wiring guidance when you had already done it in either schematics or breadboard.* or "*I think the wiring and the connection between breadboard and schematic view was great.*").

## 5 Discussion

The simulator seems that it was extremely useful when the students did not have access to the hardware and less useful when they have access to practical sessions. This result is in line with other experiences using simulators [14].

From my personal experience, I have noticed that using Fritzing and the simulator has improved the learning of the students. Using Fritzing, they can work in three different views (breadboard, schematic and PCB). As the exercises forced the students to implement the circuits in a breadboard and draw the schematics for each exercise, they learn to draw and interpret schematics. In addition, the functionality of highlighting all the terminals that are connected together (left click on a terminal), helps them to understand how breadboards, net labels and power symbols work.

While I think that working with hardware is necessary, the simulator has a couple of advantages. In some occasions, detecting a failure in hardware is not straight forward and it is easy to see this in the simulator. As an example, my students need to read the status of a switch. While the exercises tries to force them to use a pull-up or pull-down resistor, some of them solve this task by creating a short-circuit when the switch is pressed. While this solution is not acceptable, they cannot notice it as the short-circuit does not produce any noticeable effect except that there is no power (e.g., when taking 5 volts from a voltage regulator with short-circuit protection). Instead, it is easy to see that this solution is not valid in the simulator, as they can see a clear smoke coming out of the battery. These kinds of issues also appear when a circuit is working outside of its specifications.

The feedback of the students in 2021 was used to improve the simulator. However, the simulator still has some limitations. It can only simulate one circuit at a time. Several circuits need to be connected to each other to be able to simulate them in Ngspice or two independent simulations need to be performed. There are no interactive parts. The users can change the properties of the parts in the inspector pane, but they cannot drag the knob of the potentiometers. This makes the experience less realistic. And finally, the simulator only allows for DC operating point analysis. While this is enough for basic electronic teaching, an AC simulation is still missing.

## 6 Conclusion and Future Work

Teaching online electronics poses some challenges on how to replace lab sessions with real hardware. In this paper, I have described how this shortcoming can be addressed with a simulator that tries to replicate the experiences of working with real circuits. The simulator has been implemented in Fritzing and allows students to implement the circuits on a breadboard and, at the same time, draw the schematics. Results shown that students find the simulator useful, and my personal experience suggests that a simulator can still be useful even if students have access to real hardware.

In the short term, I will focus on improving the simulator: increasing the parts with spice models, developing interactive parts and adding AC simulation and oscilloscopes while maintaining ease of use. Long term efforts include integrating the simulator in a holistic robot simulator (HoRoSim) [15] or introducing game-based learning in the simulator.

**Acknowledgements** Thanks you to Kjell Morgenstern, who compiled the simulator, and to Peter Van Epp and Phil Duby, who fixed the original multimeter part. Finally, I would also express my gratitude to the students who tested the simulator and gave valuable feedback.